\documentclass[runningheads]{llncs}
\usepackage{graphicx,array}
\usepackage{amsmath,amssymb} 
\usepackage{color}
\usepackage[width=122mm,left=12mm,paperwidth=146mm,height=193mm,top=12mm,paperheight=217mm]{geometry}
\usepackage[font=small]{caption}
\usepackage{subcaption}
\begin{document}

\pagestyle{headings}
\mainmatter
\title{The Open World of Micro-Videos}

\titlerunning{The Open World of Micro-Videos}

\authorrunning{Nguyen et al.}

\author{Phuc Xuan Nguyen$^{1}$, Gregory Rogez$^{2}$, Charless Fowlkes$^{1}$, Deva Ramanan$^{3}$}
\institute{UC Irvine$^{1}$, INRIA$^{2}$, Carnegie Mellon University$^{3}$}

\maketitle

\begin{abstract}
Micro-videos are six-second videos popular on social media networks with
several unique properties. Firstly, because of the authoring process, they
contain significantly more diversity and narrative structure than existing
collections of video ``snippets''. Secondly, because they are often captured by
hand-held mobile cameras, they contain specialized viewpoints including
third-person, egocentric, and self-facing views seldom seen in traditional
produced video. Thirdly, due to to their continuous production and publication
on social networks, aggregate micro-video content contains interesting
open-world dynamics that reflects the temporal evolution of tag topics.  These
aspects make micro-videos an appealing well of visual data for developing
large-scale models for video understanding.  {\em We analyze a novel dataset of {\bf
260 thousand} micro-videos labeled with {\bf 58 thousand}  tags.}  To analyze
this data, we introduce viewpoint-specific and temporally-evolving models for
video understanding, defined over state-of-the-art motion and deep visual
features. We conclude that our dataset opens up new research opportunities for
large-scale video analysis, novel viewpoints, and open-world dynamics.
\end{abstract}

We examine an increasingly prevalent form of media known as {\em micro-videos},
time-constrained (typically 5-10 second) video clips commonly used on
social networking sites such as Instagram and Vine. Micro-videos can be
interpreted as the visual analog of a character-limited micro-blogs or
``tweets''~\cite{kwak2010twitter}.  An estimated 12 million micro-videos are
posted to Twitter each day. The number of micro-videos produced 
{\em surpasses the total inventory of YouTube every 3 months}.  From an
applied perspective, this flood of visual data is increasingly important and
has unique characteristics that are not addressed by existing computer vision
methodologies and benchmarks. We further argue that the microvideo format
offers unique opportunities for basic research in building systems that address
lifelong learning in {\em open-world} visual domains.

{\bf Ease of collection/processing:} A particularly attractive aspect of
micro-videos is the ease of large-scale collection, storage and processing.
While large-scale datasets~\cite{everingham2014pascal,deng2009imagenet} have
revolutionized static-image processing, the counterpart for video-based
recognition does not appear to exist. One reason is that it is notoriously
challenging to collect, store, and process a large diverse video collection
because of resource constraints. Indeed, existing video collections often
contain multiple snippets cropped from a few longer videos (to simplify the
collection process) ~\cite{hmdb,imagenetVid}. As we experimentally validate, in
Sec.~\ref{sec:hmdb}, this limits their diversity when compared to micro-video
collections.

\begin{figure}[ht!]
\centering
  \includegraphics[width=\linewidth]{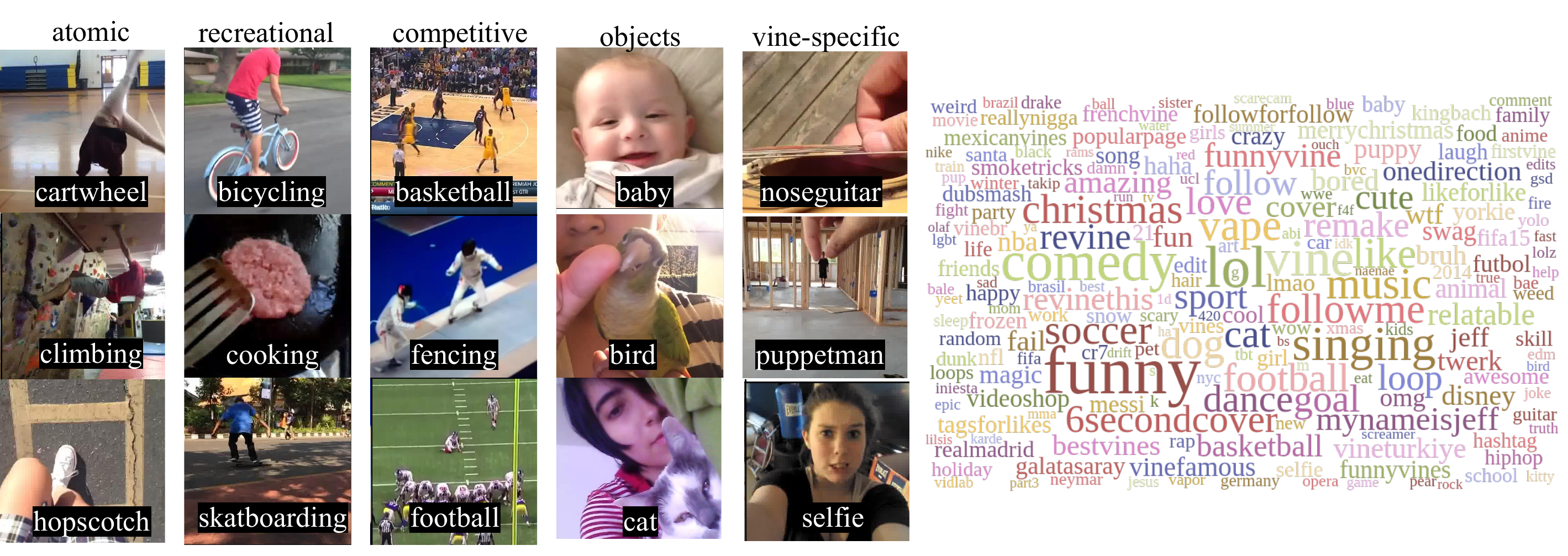}
   \caption{Micro-videos lie in a regime between images and traditional videos, encoding
 semantically rich micro-narratives while remaining tractable to collect, store
 and process.  Mobile-videographers often interact with the scene and its
 subjects resulting in a wide range of camera viewpoints including {\bf
 egocentric} views of activities and {\bf self-facing} shots where a single
 individual is both the photographer and subject.  Our dataset, {\bf MV-58k},
 includes common tags about actions and objects seen in other computer vision
 datasets as well as specific tags such as {\tt \footnotesize \#noseguitar} and
 {\tt \footnotesize \#puppetman} which are video-graphic styles unique to the
 micro-video sharing service Vine.  Distribution of videos per tag is highly
 skewed.  At the time of submission our dataset totals more than 250,000 videos
 and includes at least 1000 videos for 50\% of the tags.}  
\label{fig:splash}
\vspace{-20pt}
\end{figure}

{\bf Micro-narratives:} An intriguing aspect of micro-videos is that they
arrive ``ready-for-analysis'', due to the constraint that users are forced to
trim content to fit within the six-second restriction.  Micro-videos often
contain smaller {\em nano-shots}, each sometimes containing a unique viewpoint,
that are spliced together so as to contain a complete narrative
(Fig.~\ref{fig:shot}).  Indeed, micro-videos are now given their own categories
in established film festivals~\cite{kelly2014slow}.  As a result of this
intense degree of editing, each frame of a micro-video typically has high
information content compared to frames in a longer unconstrained video. This
changes strategies for automated understanding and may even eliminate the need
for common video preprocessing steps like keyframe or shot selection.

{\bf Viewpoint:} A unique technical aspect of micro-videos found on social
networks is camera viewpoint (Fig.~\ref{fig:view_tag}). Current action datasets
in computer vision focus on {\em third-person} depictions of actions, where a
person(s) performing an action or activity is framed in the view. In contrast,
a significant fraction of socially-driven micro-videos include {\em egocentric}
viewpoints, where the photographer is participating in the action.  Another
camera configuration is a {\em self-facing} viewpoint, or ``selfie'',  where a
single user is both the photographer and subject.  This is particularly
interesting in the study of interactions photographers and their subjects
\cite{marien2002photography}.  Indeed, the ``selfie" is commonly recognized as
a medium for embodiment and empowerment because of the unique
photographer-subject interaction it afford~\cite{jones2014eternal}.  Such
diverse camera viewpoints represent new modes of video acquisition that are not
typically addressed in previous work and deserve a closer look from the
computer vision community.

{\bf Open-world dynamics:} Micro-videos come labeled with hashtags that enable
search and play an important role in social communication.  These tags provide
a form of supervision, automatically labeling our diverse, multi-view,
pre-trimmed dataset. In contrast to existing efforts to develop top-down
ontologies for activities~\cite{caba2015activitynet} or events~\cite{Ye:MM15},
tags form an open-world vocabulary whose usage and semantics changes
dynamically over time~\cite{cunha2011analyzing}.  For example, the tag {\tt
\#trump2016} did not exist until recently, while the visual meaning of the tag
{\tt \#apple} expands whenever a new iPhone is released.  Microvideos thus provide a unique
opportunity to explore learning {\em in the open world}, where distributions of
visual semantics follow long-tail statistics that change time over time. Taking
a {\em bottom-up, data-driven} approach to video content semantics more
accurately reflects naturally-occuring long-tail distributions that typically
suppressed in hand-curated datasets.  The dataset we present is thus large, has
dynamic temporal variations and is continually growing is size. We analyze a
snapshot as of Feb 2016 consisting of 264,327 videos with 58,243 tags. Based on
a conservatice estimate of the current growth rate, the dataset will approximately include 700,000 videos and 120,000 tags by the
time of ECCV.

\begin{figure}[t!]
\centering
\begin{subfigure}[b]{0.6\linewidth}
  \resizebox{\linewidth}{!}{
  	\includegraphics{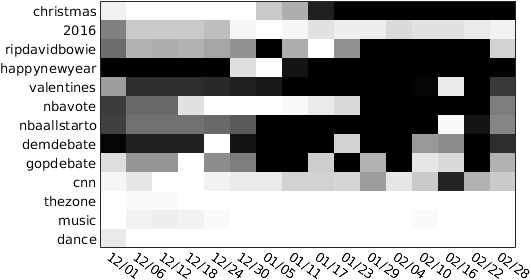}
  }
  \caption{tag usage dynamics}
\end{subfigure}~
\begin{subfigure}[b]{0.3\linewidth}
  \resizebox{\linewidth}{!}{
    \begin{tabular}{cc}
   {\large  Jan-Feb 2015} & {\large July-Aug 2015}\\
    \includegraphics[width=\linewidth]{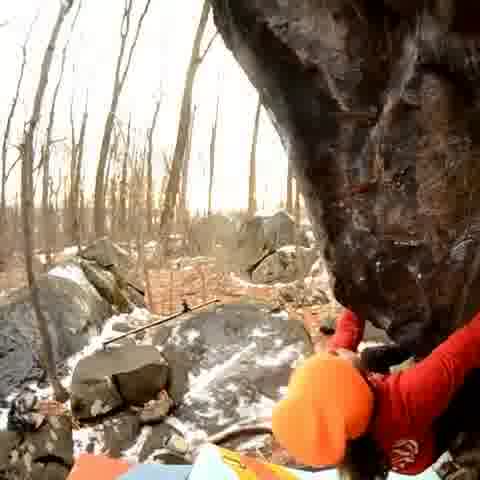}&
    \includegraphics[width=\linewidth]{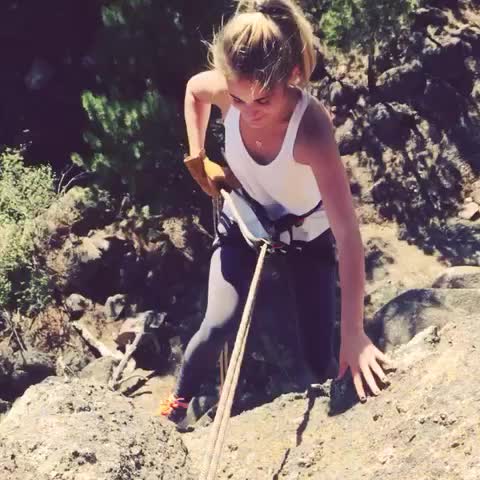}\\
    \includegraphics[width=\linewidth]{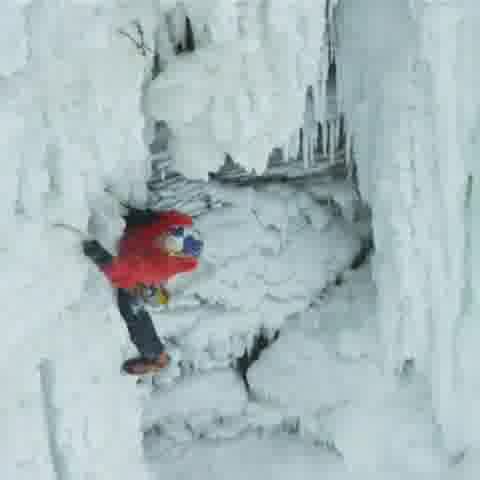}&
    \includegraphics[width=\linewidth]{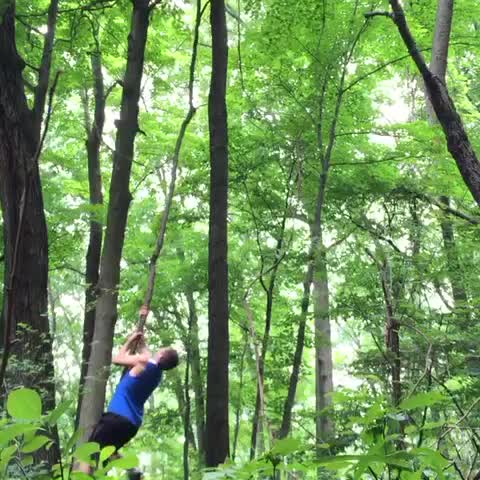}\\
    \end{tabular}
  }
  \caption{appearance dynamics}
\end{subfigure}
\vspace{-10pt}
\caption{Unique temporal structure in microvideos. We visualize changes in tag
label priors over time by plotting their popularity rank as a heatmap (brighter denotes higher popularity).  Note that many tags exhibit large fluctuations; {\tt \#gopdebate} is ranked first during the weeks
of the associated events, but dramatically decreases otherwise. We can also interaction between the tags; {\tt \#cnn} rises in popularity as {\tt \#demdebate} and {\tt \#gopdebate} become popular. Some tags ({\tt \#music}) exhibit relatively stable popularity. The distribution of visual appearance associated with some tags also exhibit temporal dynamics.  {\tt \footnotesize \#climbing} shows large temporal
variations over summer and winter months due to changes in scenery (desert
rocks versus ice) and equipment.
\label{fig:dynamics}}
\vspace{-20pt}
\end{figure}


\section{Related Work}

There is little existing work on micro-video analysis, as the medium itself is
new. Redi {\em et al.} \cite{vinecreativity} explore the problem of finding
creative micro-videos, inspired by similar studies of image quality assessment.
Sano {\em et al.}~\cite{sano2014degree} analyze the problem of detecting loop
micro-videos (that are designed to be played in a continuous six-second loop).
Here we focus on analyzing general properties of micro-videos, with the
explicit goal of constructing a new, large-scale benchmark for
temporally-evolving tag prediction.

{\bf Viewpoint modeling:} A unique contribution of our dataset is the
diversity of camera viewpoints.  Existing video benchmarks for action
recognition have focused on third-person viewpoints (e.g., HMDB~\cite{hmdb},
UCF101~\cite{soomro2012ucf101}, Hollywood-2~\cite{hollywood2},
UT-Interaction~\cite{utinteraction} and Olympic Sports~\cite{olympic}).
Wearable cameras such as Google Glass and GoPro have spurred interest in
analyzing egocentric 
views~\cite{fathi2011understanding,kitani2011fast,pirsiavash2012detecting,lee2014predicting}.
Compared to existing action and egocentric datasets, our user-generated
micro-videos contain a wider variety of categories (tags) and viewpoints (e.g.,
self-facing) with richer narrative content, even in a single clip. To
understand and highlight these differences we train mixture-of-viewpoint models
that specifically target viewpoint variations in dynamic micro-videos
(Sec.~\ref{sec:model}) and carry out an extensive comparison with
HMDB~\cite{hmdb} (Sec.~\ref{sec:exp}).

\begin{figure}[t]
    \centering
      \begin{subfigure}[b]{.35\linewidth}
      \raisebox{1.2\height}{
	  \begin{tabular}{ccc}
		Direct & Indirect & Nonvisual\\
        \includegraphics[width=.31\linewidth]{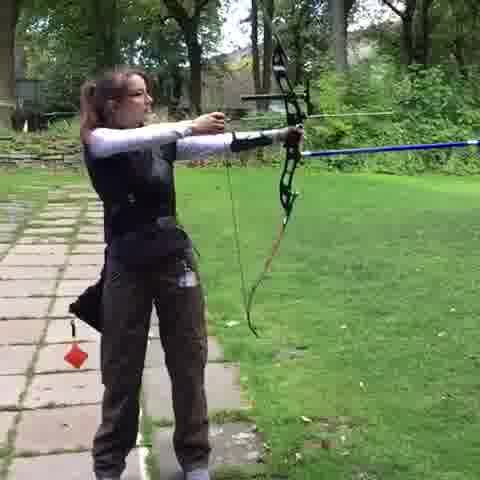}&
        \includegraphics[width=.31\linewidth]{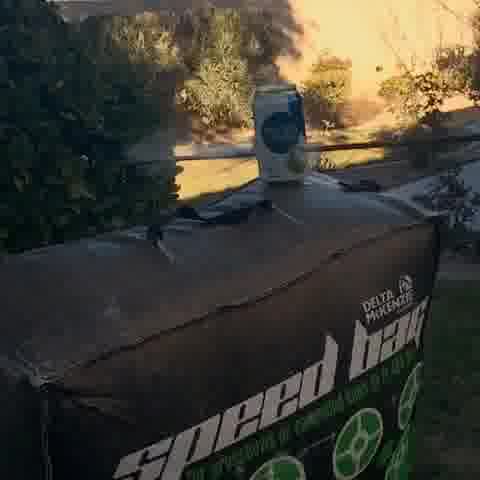}&
        \includegraphics[width=.31\linewidth]{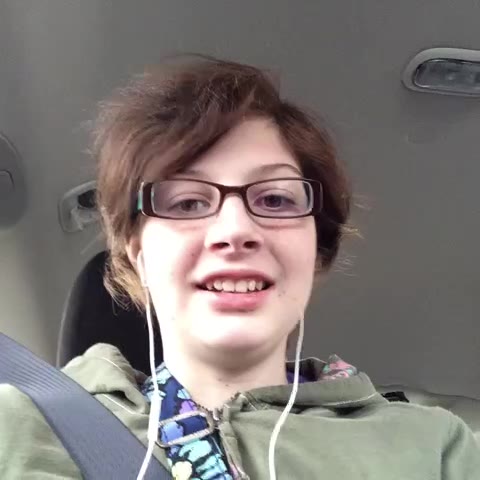}
	  \end{tabular}}
		\vspace{-5pt}
        \caption{\#archery videos}
      \end{subfigure}
      ~
      \begin{subfigure}[b]{.6\linewidth}
        \includegraphics[width=\linewidth]{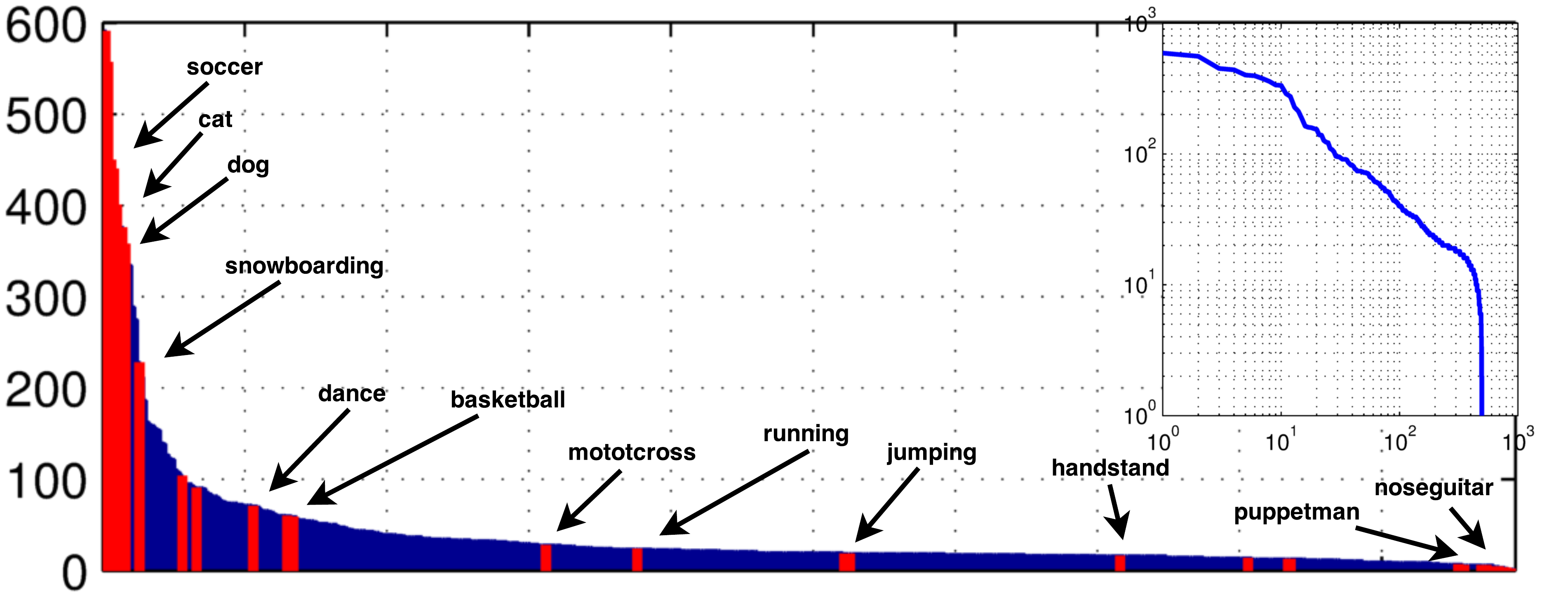}
\vspace{-5pt}
        \caption{Long-tail distribution of tags}
      \end{subfigure}
    \vspace{-3mm}
    \caption{
    (a) Videos tagged with {\tt \#archery} may show direct evidence of the
    tagged activity, circumstantial evidence (video of a can which is suddenly
    pierced by an arrow), or non-visual evidence (a girl talking about an
    archery competition).  For diagnostic purposes we construct a hand-curated
    dataset (MV-40) containing only videos with direct visual evidence for a
    subset of 40 tags selected to span both common and rare tags, as shown in
    the long-tailed distribution (b).  Inset shows the same distribution on a
    log-log scale.}
    \label{fig:visbreakdown}
\vspace{-20pt}
\end{figure}

{\bf Closed vs open vocabularies:} Traditionally, video datasets in computer
vision have been labeled with a fixed ontology of activities or
events~\cite{caba2015activitynet,Ye:MM15,karpathy2014large}. An alternative
perspective (popular in the multimedia community) is to formulate the problem
as a multi-label tag or concept prediction
task~\cite{vahdat2013handling,yang2011discriminative,aradhye2009video2text,toderici2010finding,ulges2009tubetagger}.
Our dataset falls into this later camp. In terms of size and diversity, the
most relevant prior work appears to be Sports-1M~\cite{karpathy2014large}
which contains 1M videos in 487 categories, and EventNet~\cite{Ye:MM15}, which
contains 95K videos labeled with 5K concepts. Our dataset already includes 2X
more videos and 10X more concept tags. Unlike other video datasets, our data
also includes timestamps which allow us to study temporally-varying semantics,
a relatively unexplored concept in vision, with the notable exception
of~\cite{kim2010modeling}.  Importantly, tag frequency distributions are highly
imbalanced, following a natural long-tail distribution
(Fig.~\ref{fig:visbreakdown}). While highly imbalanced class distributions are
somewhat uncommon in current vision datasets, they appear to be a fundamental
aspect of life-long learning in the open-world~\cite{chen2013neil}.  With the
advent of deep architectures that appear capable of transferring knowledge
across imbalanced classes~\cite{bengio2011expressive}, we think the time is
right to (re)consider learning in the open-world!

\section{Dataset}
\label{sec:data}

In this section, we describe our (continually-running) data-collection process
and analyze the statistics of micro-video tags, shots and views that make our
dataset distinct from existing video benchmarks.

{\bf Streaming dataset collection:} We collect a stream of Vine videos by daily
querying of Vine's API~\cite{vineapi}. To ensure a diverse stream, we query the
300 most popular and 300 most recent videos across multiple community-curated
channels (Comedy, Sports, Musics). We typically obtain ~6000 videos daily. Each video is associated with a collection of hashtags
that are added to our open vocabulary. On average, a video contains 1.56 hashtags, but this statistic is skewed by the fact that
nearly half the videos do not contain any tags (56\%).This indirectly motivates one practical application of our dataset -
automatic prediction of hashtags. We attempted to automatically merge similar
tags through linguistic normalization, but found this merging did little to
change our label space (perhaps because users have an incentive to used
normalized hashtags that are already searchable). We visualize our overall
distribution of tags in Fig.~\ref{fig:splash}. We use {\bf MV-58k} to refer
to a snapshot of this datastream collected during the period Dec-2015 to Feb-2016.  We provide additional statistics of this data collection and
projections in the supplemental materials.

\begin{figure}[t!]
  \centering
  \begin{minipage}{0.7\linewidth}
    \begin{subfigure}[t]{\linewidth}
      \includegraphics[width=\linewidth]{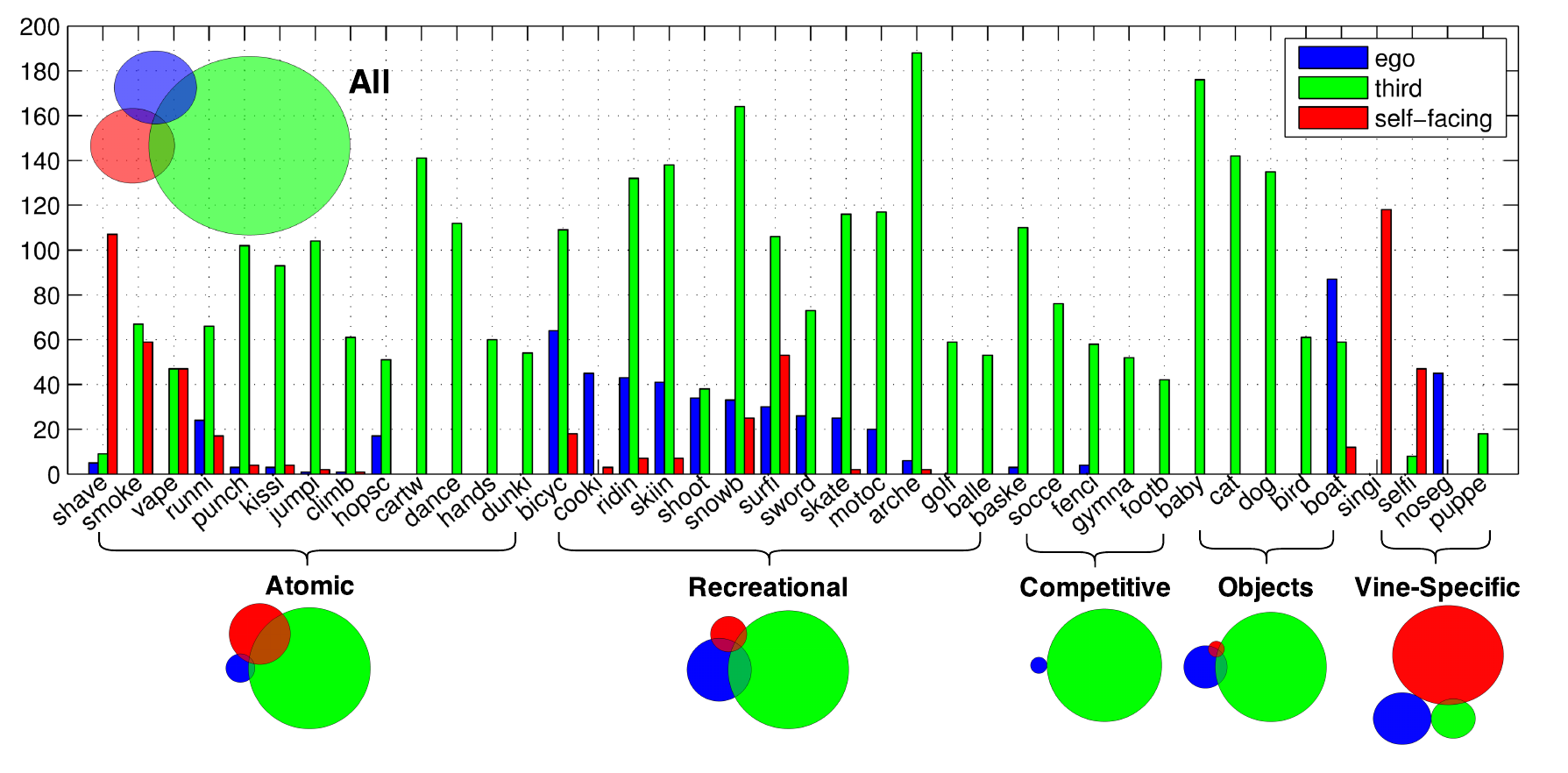}
    \end{subfigure}
  \end{minipage}
  \begin{minipage}{0.27\linewidth}
    \begin{subfigure}[t]{\linewidth}
      \resizebox{\linewidth}{!}{
         \begin{tabular}{ccc}
         & {\tt \Large \#surfing} & {\tt \Large \#bicycling}\\
         \rotatebox[origin=l]{90}{\hspace{30pt} \Large Third}
         & \includegraphics[width=1\linewidth]{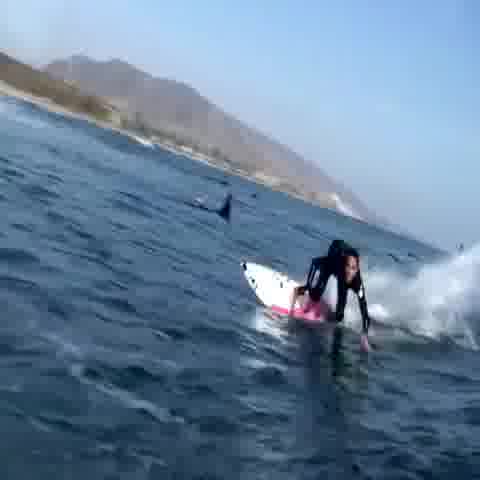}
         & \includegraphics[width=1\linewidth]{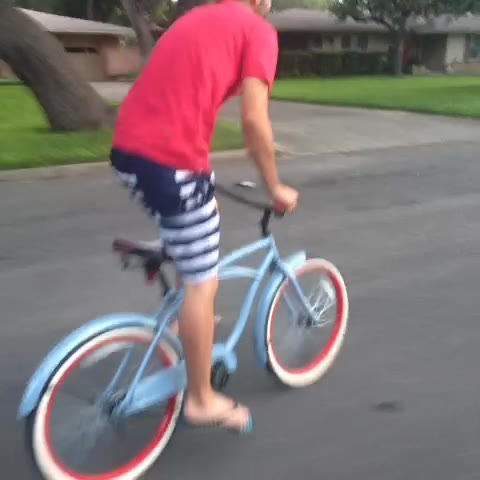} \\
         \rotatebox[origin=l]{90}{\hspace{30pt} \Large Ego}
         & \includegraphics[width=1\linewidth]{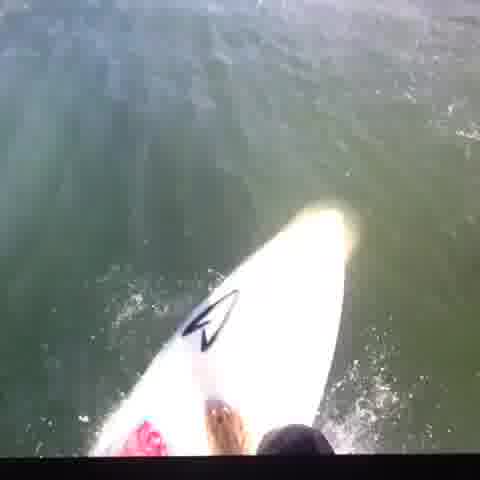}
         & \includegraphics[width=1\linewidth]{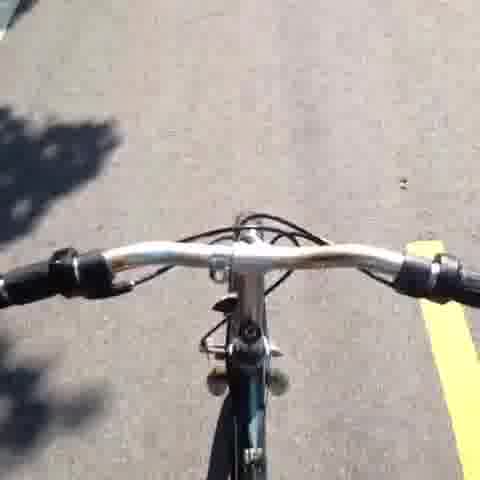} \\
         \rotatebox[origin=l]{90}{\hspace{30pt} \Large Self}
         & \includegraphics[width=1\linewidth]{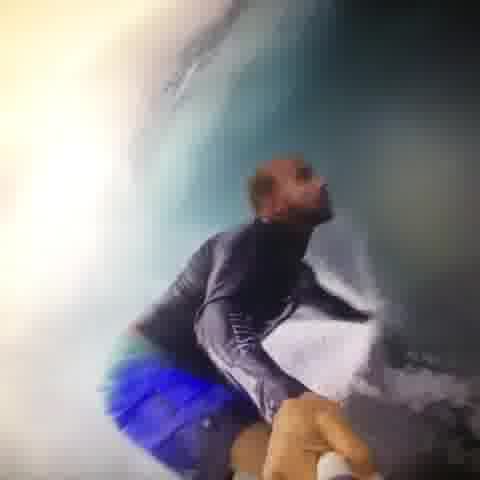}
         & \includegraphics[width=1\linewidth]{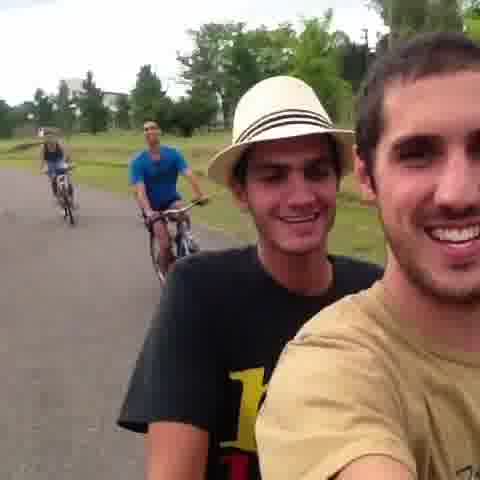} 
         \end{tabular}
      }
   \end{subfigure}
 \end{minipage}
    \caption{The unique viewpoints of microvideos. ({\bf a}) shows the
    distribution of camera viewpoints annotated in MV-40.  Most tags tend to be
    associated with third-person viewpoints. Some atomic actions such as {\tt
    \footnotesize \#shave}, {\tt \footnotesize \#smoke}, and {\tt \footnotesize
    \#vape} (smoking with an electronic cigarette) are captured with
    self-facing views. Tags for recreational sports often include egocentric
    views (where the photographer is engaged in the action), but this is less
    likely for competitive sports (since it may be difficult to hold the
    camera). Many Vine-specific tags make use of self-facing viewpoints. From
    the Venn diagram embedded in the bar graph, we can see that there's a
    significant amount of videos that contain more than one viewpoint and some
    has all 3 viewpoints. We show examples of frames with unique viewpoints in
    {\bf (b)}. The biking frame could be labeled as both {\em ego} and {\em
    third} because the photographer and subjects are engaged in a ``social''
    activity. We also show a {\em self-facing} viewpoint made possible through
    a specialized camera mount.}
    \label{fig:view_tag}
\vspace{-20pt}
\end{figure}

{\bf Curation:} To examine the amount of ``noise'' in the tag stream and
perform diagnostic comparisons to existing activity datasets, we manually
curated a subset of 40 tags and their associated 4000 videos.  We selected 40
representative tags that span both common actions as well rare tags
``in-the-tail''. We ``clean'' this dataset by merging synonymous tags (e.g.,
{\tt \small \#horseback} and {\tt \small \#horsebackriding}), removing
mistagged (spam) videos, and removing videos in which there was only
circumstantial visual evidence for the tagged activity (see
Fig.~\ref{fig:visbreakdown} for an example). We added additional
annotations to each video including viewpoint and a dominant (tag) category. The latter allows us to recast tag prediction as a K-way classification problem, simplifying our diagnostic analysis.
We refer to this curated dataset as MV-40, and contrast this with (the fixed
snapshot of) our uncurated, open-world dataset MV-58K. We organize MV-40 into
broad categories of atomic actions, recreational activities, competitive
sports, objects, and vine-specific.  We visualize our two-level taxonomy and
provide visual examples in Fig.~\ref{fig:splash}.

{\bf Long-tail distributions:} Our collection process reveals a salient
property of open-world microvideos; they follow long-tail distributions of
tags. This significantly complicates learning because there will be some tags
for which we have little training data. While traditionally a notorious
challenge for machine learning, our analysis suggests that hierarchical feature
learning (with CNNs) can learn to share, or transfer knowledge from the
data-rich tags to the data-sparse tags (i.e., {\em one-shot learning}). For
example, even if we have few examples of {\tt \#dunking}, mid-level features
learned for a data-rich tag such as {\tt \#basketball} may still be useful for
the former class.

\begin{figure}[t!]
    \centering
    \includegraphics[width=.7\linewidth]{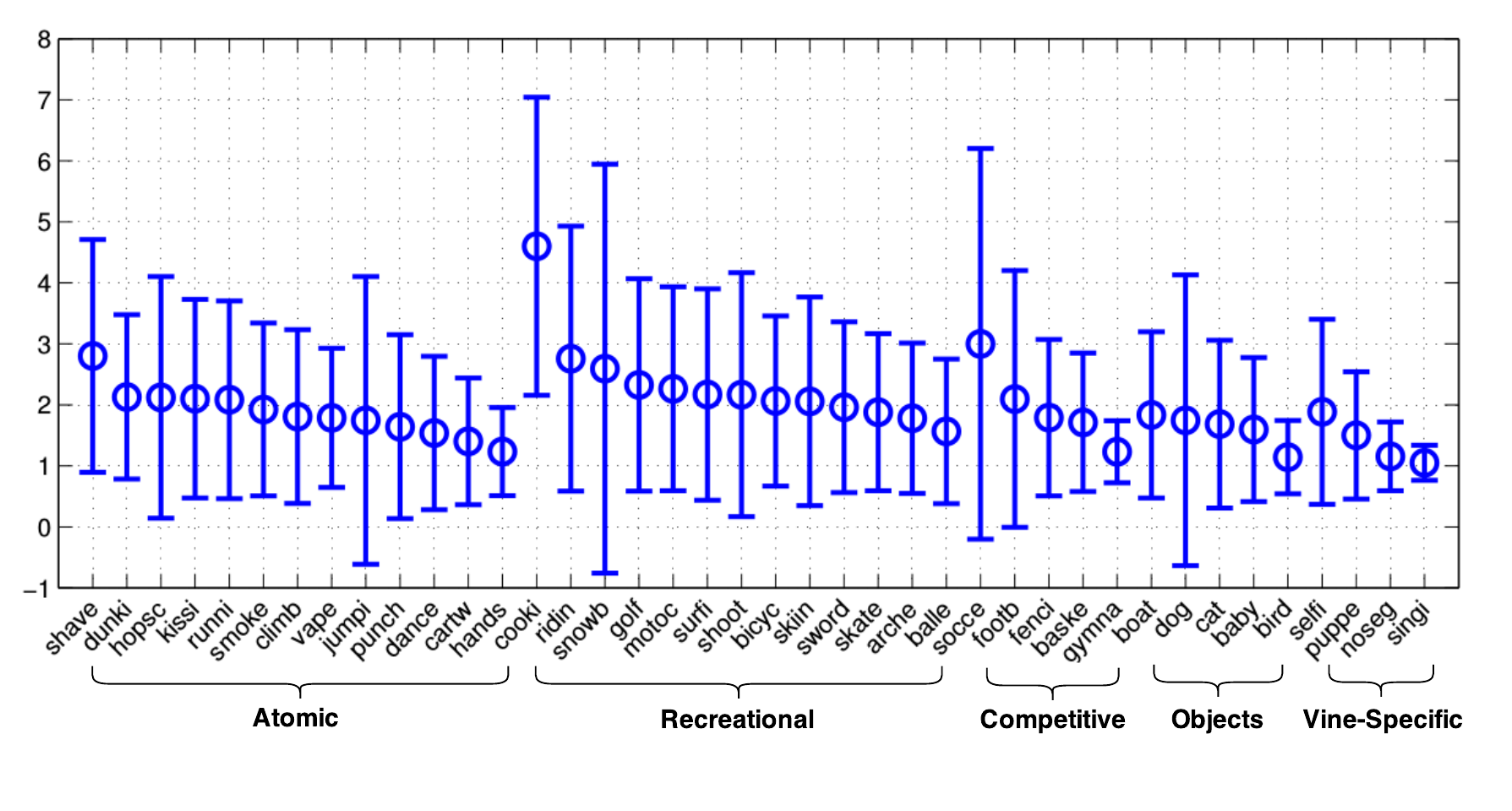}
\vspace{-20pt}
    \caption{Unique shot structure in microvideos. We plot the mean and variance of the
  number of shots in a video.  Over 34\% of our videos contain more than a
  single shot. A small percentage(3.75\%) contain 6 or more shots, implying
  that many shots are less than 1 second in duration. In terms of per-tag
  statistics,  {\tt \footnotesize \#cook}ing contains a large number of spliced
  shots, necessary to temporally compress and summarize such a long duration
  activity.  {\tt \footnotesize \#snowboard}ing and  {\tt \footnotesize
  \# soccer} have a large variance because they contain some compilation videos
  composed of many shots.
}
    \label{fig:shot}
\vspace{-20pt}
\end{figure}

{\bf Temporal dynamics:}  Our open-world dataset collection has another notable
property - both the frequency of tag usage and the visual appearance semantics
associated with a given tag evolve over time.  In probabilistic terms, we can
interpret such dynamics and changes in the {\em prior} of labels
(Fig.~\ref{fig:dynamics}) and the {\em likelihood} of image features
conditioned on the tag label (Fig.~\ref{fig:time}). This suggests developing
approaches for continually retraining models as new data becomes available, an
idea we explore in Sec.\ref{sec:exp}. An extreme case arises when a new
tag first appears in the data stream there are no training examples available
(e.g., {\tt \#trump2016} did not exist until recently).  In our current
experiments, we simply fail to predict such tags at test time. However, we
point out that the temporal appearance of new tags provides a compelling
natural example of {\em zero-shot learning} where side information about the
semantic relation between tags could readily be exploited.

{\bf Shot statistics:} Micro-videos have a unique shot structure due to the
capturing mechanics and limited time constraint imposed by the app. A
Vine or Instagram user creates videos by holding a button to capture and
releasing the button to pause at any time.  He/she can later resume capturing
again until the content time limit is exceeded (6 seconds for Vine). This
interface allows and even encourages splicing together of many shots within a
single time-limited video. We plot a distribution shot length and frequency for
various tags in Fig.~\ref{fig:shot}. Some tags, such as {\tt \small \#cook}ing
tend to consistently involve a large number of shots. Indeed, we find 3.75\% of
videos have more than 6 shots, implying many shots are less than a second in
length.

{\bf Viewpoint:} To analyze the effect of viewpoint on recognition accuracy, we
manually annotate MV-40 with the viewpoint of each video as egocentric,
third-person, or self-facing.  In some cases, the viewpoint changed between
shots in the video in which case we record all the viewpoints present, as well
as the dominant one. With this annotation we can treat viewpoint prediction as
either a multi-label attribute prediction problem (where multiple viewpoints
can be present in video) or a multi-class problem (where the goal is to predict
the dominant viewpoint).  We report the per-tag viewpoint statistics in
Fig.~\ref{fig:view_tag}. Interestingly, even at the shot-level, we sometimes
find ambiguities in viewpoint. Some social activities (such as {\tt \small
\#bicycling}) involve both the photographer and subjects in view, suggesting a
simultaneous egocentric and third-person viewpoint.


\section{Methods}
\label{sec:model}
In this section, we describe several baseline models for tag prediction and
viewpoint classification. As the focus of our work is not on video feature
extraction, we make use of standard feature sets. Recent results from the
THUMOS evaluation benchmark~\cite{gorban2015thumos} suggest that CNN spatial
features (VGG~\cite{simonyan2014very}) combined with temporal motion features (IDT~\cite{idt}) make for a
reasonable video descriptor. We briefly outline our feature descriptor and
associated classification engines here.

\label{sub:Features}
{\bf Appearance features:} Recent work has shown that Convolutional Neural
Networks (CNN) produce quite effective visual descriptors for recognition
tasks~\cite{razavian2014cnn,simonyan2014very} including video analysis
\cite{ryoo2014pooled,simonyan2014two,karpathy2014large,xuCNNDiscriminative}.
We experimented with many open-source implementations of CNN architectures for 
video-feature extractors, and found that while many were effective, some placed
significant demands on processing time and descriptor storage. We refer the
reader to our supplementary material for a detailed description of our
diagnostic experiments. We found a good tradeoff in speed, storage, and
accuracy with the following simple pipeline: given a video, (1) run
off-the-shelf VGG-16 models~\cite{simonyan2014very} on 15 equally-spaced frames
from a video, (2) for each frame, extract (6144-dimensional) multi-scale
features across multiple layers~\cite{yang2015multi}, and (4) max-pool the
resulting features across the 15 frames~\cite{ryoo2014pooled}.  When compared to
standard single-frame CNN feature extraction (that extract 4096-dimensional
features~\cite{simonyan2014very}), our final video pipeline was only 15x slower
and 2x larger in storage costs.

{\bf Motion features:} Recent state-of-the-art results on video
datasets have made use of trajectory-based motion features.  We include such
features in our analysis, focusing on Improved Dense Track (IDT)~\cite{idt}.
This method is based on a ``interest-track'' framework (rather than a
space-time interest-point approach) in which short-term tracks are found from
tracking interest points across frames.  One then extracts various features
aligned to these temporal tracks, including oriented gradient histograms and
optical flow.  We quantize these descriptors based on established guidelines
for constructing a codebook, using $K=4000$ codebook entries found with
K-means.

{\bf Cue-combination:} We train tag classifiers that aggregate features by
combining multiple kernels. We experimented extensively with various feature
encodings and kernel combinations before settling on the following strategy.
We compute the similarity between two video clips $x_i$ and $x_j$ by averaging
their appearance and motion-feature similarities:
\[
K(x_{i},x_{j})=\frac{1}{2}(K_{m}(x_{i},x_{j})+K_{a}(x_{i},x_{j}))
\]
We define the motion similarity with a $\chi^2$-RBF kernel:
\[
K_m(x_{i},x_{j})=exp\left(-\frac{1}{L}\sum_{c=1}^{L}\frac{d(x_{i}^{c},x_{j}^{c})}{A^{c}}\right), \quad x^c_i,x^c_j \in {\mathcal R}^{4000}
\]
\noindent where $A_c$ is the average $\chi^2$ distance between all videos in
the training data, L is the number of motion channels (Traj, HOG, HOF, MBHx,
MBHy), and $d(x_{i}^{c},x_{j}^c)$ is the $\chi^2$ distance between $x_i$ and
$x_j$ with respect to the $c$-th channel.  We measure the appearance similarity
between two clips using a linear kernel:
$K_{a}(x_{i},x_{j})=\sum_{f=1}^{N}(x_{i}^{f} \cdot x_{j}^{f})$ summed over the
$N=15$ static feature channels extracted by the CNN.  Given a training
vocabulary of $K$ tags, we train $K$ binary (one-vs-all) kernelized SVMs using
the LIBSVM package ~\cite{chang2011libsvm}.  Finally, we calibrate each
predictor using Platt scaling~\cite{platt1999probabilistic}.

{\bf Viewpoint mixtures:} Different viewpoints of the same tag can have
significant differences in video content, as shown in Fig.~\ref{fig:splash}.
For example, a third-person and egocentric {\tt \small \#bicycling} video
contain very different motions and appearances. To analyze such variations in
the MV-40 diagnostic dataset, we train viewpoint-specific models for each tag.
The final confidence associated with a tag prediction is the maximum score
across the three viewpoint-specific models.

{\bf Temporally-adapted models:} To explore the temporal evolution of tag semantics and
videos, we evaluate models trained with videos sampled over different temporal
windows.  Consider the task of predicting the label $y$ for a video $x$
collected at time $t_0$ using with models trained on a stream of timestamped
training videos indexed by time.  We consider models trained on three different
subsets of training data:
\begin{align}
  \{x_t,y_t: \forall t\} &\rightarrow \text{non-causal model}\\
  \{x_t,y_t: t < t_0\} &\rightarrow \text{causal model}\\
  \{x_t,y_t:  t_0 - \Delta < t < t_0\} &\rightarrow \text{adaptive model}
\end{align}
where $\Delta$ is a specified window size. The above approach can be simplified by making some assumptions about the nature of temporal variation. For example, if we assume that the popularity of tags changes over time but their appearance models do not, one can model efficiently model temporal dynamics with a {\em statistical prior shift}~\cite{saerens2002adjusting}. Intuitively, dynamics can be captured with a fixed set of posterior class predictions that are reweighted by dynamically-varying tag priors. Unfortunately, this requires access to tag priors on test data from the future, which violates causality. Instead, we assume that tag priors vary smoothly over time, and simply use a weighted estimate of recent tags' popularity. We found that the simple approach of applying temporally-weighted Platt rescaling (using a weighted dataset where recent videos are given more importance) outperformed an explicit prior model.


\section{Experiments}
\label{sec:exp}
In this section, we present an extensive set of experiments on our dataset
and refer the reader to the supplementary materials for additional tables
and figures.  We focus on three sets of experiments: a diagnostic evaluation of
features and viewpoints on our curated dataset (MV-40), its relation to popular benchmarks such as HMDB~\cite{hmdb}, and analysis of the open-world dynamics of MV-58K.

\subsection{Diagnostics}
First, we analyze various aspects of our dataset and recognition pipeline,
focusing on the curated and annotated MV-40 subset.

\begin{figure}[t!]
  \scriptsize
  \centering
    \begin{subfigure}[b]{0.4\linewidth}
      \begin{tabular}{lccc}
        & IDT & CNN & IDT+CNN\\
      \hline
      All      & 53.90 & 62.13 & 68.82\\
      \hline
      Atomic   & 47.72 & 46.58 & 57.19\\
      Recreational & 60.81 & 69.09 & 76.06\\
      Competitive  & 49.86 & 60.29 & 64.64\\
      Objects  & 46.14 & 70.37 & 73.47\\
      Vine-Specific & 63.25 & 65.81 & 68.38\\
      \hline
      Ego      & 63.25 & 73.15 & 79.49\\
      Third    & 51.45 & 60.00 & 67.14\\
      Self     & 59.27 & 61.82 & 68.40\\
      \hline
      \end{tabular}
      \caption{Feature performance}
   \end{subfigure} 
  ~
    \begin{subfigure}[b]{.3\linewidth}
      \begin{tabular}{lccc}
        Feature & Pos & Neu & Neg\\
        \hline
        IDT & 62.65 & 64.98 & 64.34\\
        CNN & 58.28 & 60.82 & 60.47\\
        IDT+CNN & 68.57 & 70.61 & 70.40\\
        \hline
      \end{tabular}
      \caption{View-specific mixtures}
    \end{subfigure}
  ~
    \begin{subfigure}[b]{.25\linewidth}
        \includegraphics[width=.8\linewidth]{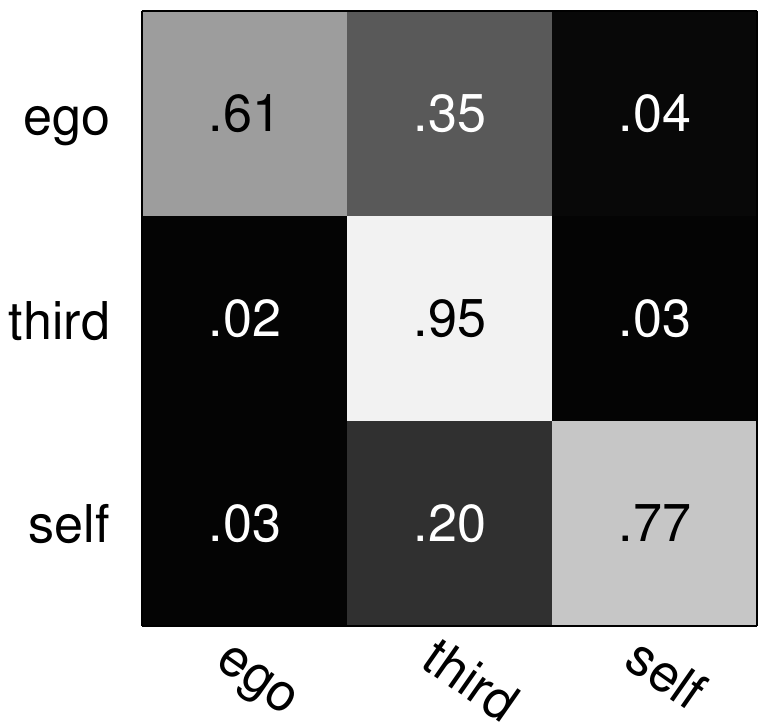}
        \caption{Viewpoint confusion}
    \end{subfigure}
	\vspace{-5pt}
  \caption{(a) Performance of different feature sets on MV-40 broken down by
  tag type and viewpoint. (b) plots the perfomance of view-point specific tag models, exploring different choices of positive and negative data. (c) visualizes the 3-way class confusion matrix for viewpoint prediction. See text for more details.}
  \label{fig:feat}
\vspace{-15pt}
\end{figure}

{\bf Feature comparisons:} We begin by comparing the performance of various
combinations of our features in Fig.~\ref{fig:feat}-(a). We observe CNN features
outperform IDT in most category groups except `Atomic'. Trajectory-based motion
and appearance-based deep features are particularly effective when combined,
indicating that they capture complementary cues.  Supplementary materials
include the class confusion matrix over all 40 tags for
the IDT+CNN feature combination.

{\bf View-specific mixtures:} We next evaluate the performance of view-specific
tag classifiers and compare to results with a single
classifier per tag.  To ensure sufficient training data for each mixture
component, we only train a mixture for a specific view if there are more than
20 videos in that viewpoint (11 classes out of 40 satisfy this criteria in
MV-40).  When training, one can treat clips from the same tag but different
viewpoints as positive training examples ({\bf Pos}), negative training
examples ({\bf Neg}), or such clips can be treated as neutral and ignored ({\bf
Neu}). Fig~\ref{fig:feat}-(b) summarizes of the performance. When averaged over
all tags, the performance increase from view-specific mixtures is rather
negligible (~0.6\% for our combined features).  We also evaluate only those 11
classes for which additional views were trained.  We see a small but definite
improvement of 2.1\%. Ignoring clips from other viewpoints ({\bf Neu}) slightly
increases performance.

{\bf Viewpoint prediction:} We also investigate the task of viewpoint
prediction: what is the viewpoint of a test video? Fig~\ref{fig:feat}-(c) summarizes viewpoint confusions. Egocentric views are often
confused with third-person and the accuracies for egocentric drops
significantly.  This is consistent with Fig.~\ref{fig:view_tag}, which suggests many recreational activities involve both the
photographer and subjects involved in the action. If we score viewpoint prediction as a multilabel problem (where each video could be labeled with more than one viewpoint), accuracy for {\tt ego}, {\tt third}, and {\tt self-facing} jump to 92\%, 90\%, and 93\%. This suggests the presence of any given viewpoint can be accurately predicted.

\subsection{Comparison to existing benchmarks}
\label{sec:hmdb}
We perform an extensive comparison of our data with a
popular action recognition benchmark, HMDB~\cite{hmdb}. We use a subset of 15
tags ({\bf vine-15}) that overlap with HMDB categories and evaluate the IDT+CNN
based predictor.  Overall, the average accuracy on {\bf vine-15} is lower than
HMDB (65.99\% vs. 71.27\%), suggesting that our data is more challenging.
Torralba and Efros~\cite{torralba2011unbiased} suggest that the `performance
drop' provides a way to quantify how biased or general a dataset may be.  The
drop for models trained on Vine data is 13\%, while the drop for models trained
with HMDB is 26.75\%.  

\begin{figure}[t!]
\centering
\begin{subfigure}[t]{0.44\linewidth}
\renewcommand{\tabcolsep}{1pt}
\begin{tabular}{cc}
\includegraphics[width=.48\linewidth]{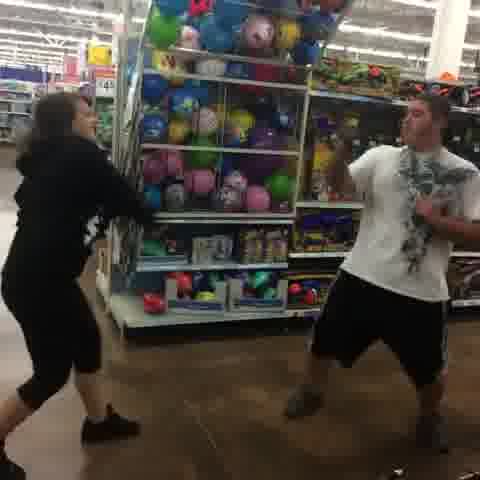}&
\includegraphics[width=.48\linewidth]{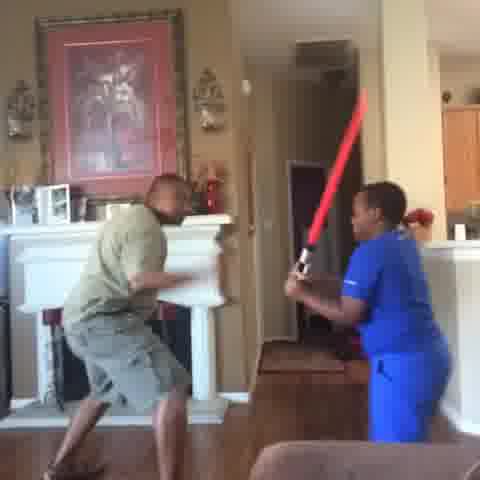}
\end{tabular}
\vspace{-3mm}
\caption{Vine swordfight}
\end{subfigure}
~
\renewcommand{\tabcolsep}{1pt}
\begin{subfigure}[t]{0.44\linewidth}
\begin{tabular}{cc}
\includegraphics[width=.48\linewidth,height=.5\linewidth]{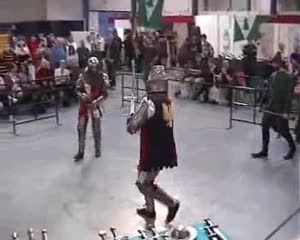}
\includegraphics[width=.48\linewidth,height=.5\linewidth]{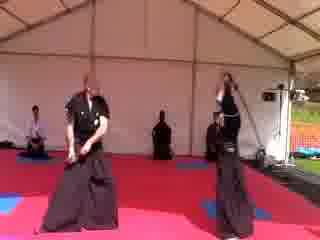}
\end{tabular}
\vspace{-3mm}
\caption{HMDB swordfight}
\end{subfigure}
~
\begin{subfigure}[t]{0.45\linewidth}
\centering
    \includegraphics[height=.95\linewidth]{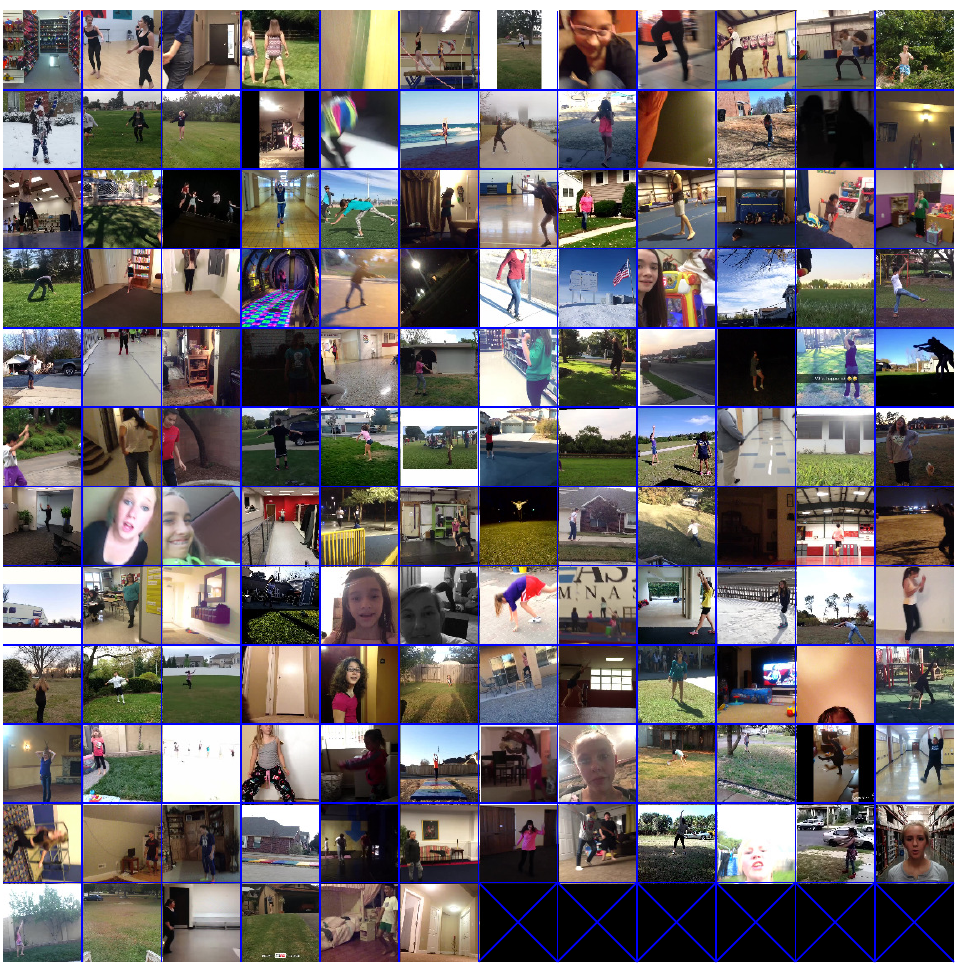}
    \caption{Vine}
\end{subfigure}
~
\begin{subfigure}[t]{0.45\linewidth}
	\centering
    \includegraphics[height=.95\linewidth]{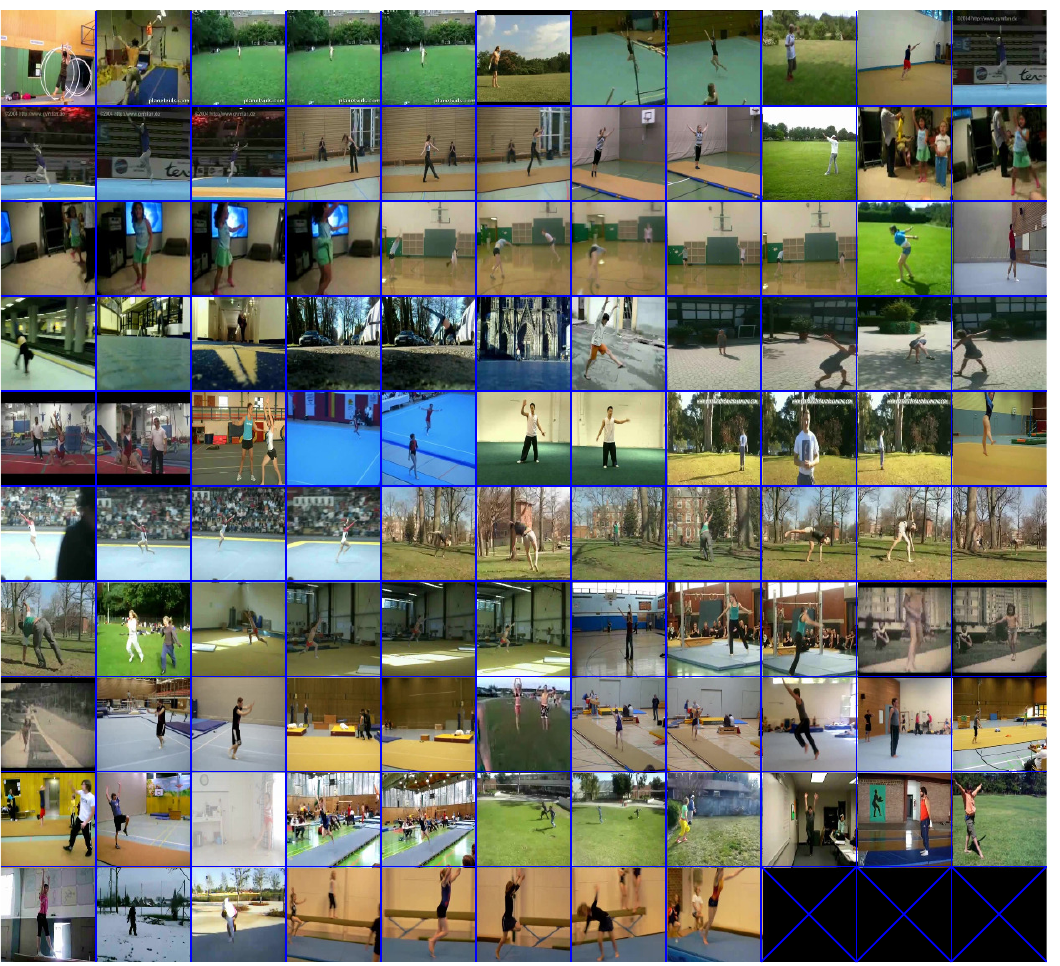}
    \caption{HMDB}
\end{subfigure}
~
\renewcommand{\tabcolsep}{3pt}
\begin{subfigure}[t]{\linewidth}
\resizebox{\linewidth}{!}{
\scriptsize
\begin{tabular}{ccc}
  \begin{tabular}[b]{|cc|c|}
  \hline Train & Test & avg \\ \hline
  vine-15 & vine-15 & 65.99 \\
  vine-15 & hmdb & 52.86 \\ \hline
  vine-3rd & vine-3rd & 65.82 \\
  vine-3rd & hmdb & 52.94 \\ \hline
  \end{tabular}
 & 
  \begin{tabular}[b]{|cc|c|}
  \hline Train & Test & avg \\ \hline
  vine-icon & vine-icon & 68.03 \\		                                                       
  vine-icon & hmdb & 53.97\\ \hline
  vine-3rd-icon & vine-3rd-icon & 68.65	 \\
  vine-3rd-icon & hmdb & 55.79 \\ \hline
  \end{tabular}
&
  \begin{tabular}[b]{|cc|c|}
  \hline Train & Test & avg \\ \hline
       hmdb & hmdb & 71.27\\                                                                                                 
       hmdb & vine-15 & 44.52\\
       hmdb & vine-3rd & 46.87\\
       hmdb & vine-icon & 45.92\\
       hmdb & vine-3rd-icon & 49.38 \\ \hline
  \end{tabular}\\
(e) & (f) & (g)
\end{tabular}}
\end{subfigure}
\caption{Qualitative difference between HMDB and Vine. The {\bf (top)} row
shows frames from the  {\tt \footnotesize \#swordfight} videos for {\bf(a)} Vine and {\bf(b)}
HMDB. Micro-videos often contain people playfully jousting in everyday scenes,
while HMDB contains more formal, staged fights and sparring events. {\bf
(Bottom)} Snapshots of {\tt \footnotesize \#cartwheel} videos for {\bf (c)} Vine and {\bf (d)}
HMDB. Multiple clips from HMDB can come from one source video, but Vine videos
are naturally collected from different sources. We evaluate various cross-dataset experiments between subsets of our dataset and HMDB in (e) - (g). \vspace{-5mm}}
\label{fig:hmdb}
\end{figure}

{\bf Viewpoint:} One might hypothesize that since HMDB contains mostly third-person views, it
won't generalize to the other viewpoint in our data.  To test this, we extract
a smaller subset {\bf vine-3rd} containing only third-person viewpoints.
Fig~\ref{fig:hmdb}-(g) shows that {\bf vine-3rd} is more similar to
HMDB, as models trained on HMDB data perform better on {\bf vine-3} than {\bf
vine-15}.
However, performance drop for models trained on {\bf
vine-3rd} is still significantly smaller than those trained on HMDB (12.88\%
vs. 24.4\%). This suggests that even accounting for viewpoint, our videos still
generalize better than HMDB.

{\bf Temporally iconic videos:} The concept of ``iconic views'' in object
recognition refers to ``easy'' images with a clear and distinctive depiction of
an object, often close cropped or in an uncluttered setting without occlusion.
We apply this notion to video, defining a
{\em temporally iconic depiction} of a tag as one where temporal clutter has
been removed by trimming down the video clip to focus on the core action.  
Temporal cropping is relevant even in micro-videos, which often contain
additional frames and shots surrounding those described by the tag.  We
manually segment each video in {\bf vine-15} to derive an iconic version {\bf
vine-icon} and use {\bf vine-3rd-icon} to denote the third-person subset. Fig.~\ref{fig:hmdb} reveals that {\bf vine-icon} and {\bf
vine-3rd-icon} are slightly easier than vine, and more similar to HMDB
(following our previous analysis).  However, the cross-dataset performance drop
for models trained on {\bf vine-icon} and {\bf vine-3rd-icon} (14.06\% and
12.86\% respectively) are still significantly smaller than the drop by models
trained {\bf HMDB} data (25.35\% and 21.89\%).  {\em To summarize, even though
third-person and temporal iconic-ness accounts for much of the difference
between HMDB and our dataset, our micro-videos can still generalize better.}

{\bf Qualitative differences:} To better understand the differences, we visualize example videos in Fig.~\ref{fig:hmdb}. {\tt \small
\#swordfight}ing in {\bf vine-15} tends to involve people playfully jousting in
social everyday scenes, while HMDB clips tend to involve formally sparring or
staged/scripted fights. This difference helps explain the why HMDB's {\tt
\small \#swordfight}s don't generalize to Vine and is consistent with the
notion that models trained from purely iconic images of objects often
generalize poorly~\cite{torralba2011unbiased}.  In the same figure, we also
plot sample frames from videos with a particular tag. {\em Perhaps surprisingly, many distinct clips in the dataset actually come from the same longer video.} This
reduces the amount of diversity in the dataset, and reinforces one of our
motivations: it {\em is} surprisingly hard to collect diverse video clips. This phenomena is not limited to HMDB, and also appears in other benchmarks such as ImageNet Video Challenge~\cite{imagenetVid}. Our
micro-video dataset, however, is inherently {\em diverse by construction} due to its dynamic and pre-trimmed nature.

\subsection{Open-World Dynamics}
In this section, we analyze properties of our open-world MV-58K dataset. One immediate issue is that open-world tags are more naturally treated as multi-label tasks (since videos can be naturally labeled with many tags). We adopt two scoring criteria from the literature on multi-label classification~\cite{makadia2008new}. $mAP_T$ treats each tag as a individual binary prediction problem, and computes the average precision (AP) of each prediction task, returning the mean AP over all tags. This approach equally weights popular and infrequent tags, and is analogous with standard mAP measures used in object detection. Alternative, $mAP_I$ first computes an image-specific AP by ranking all tag predictions specific to that image, and then returns the mean AP over all images. This latter scheme places more importance on frequent tags, and also requires that confidences across different tag predictors be calibrated. By default, we use $mAP_I$ unless otherwise specified.

\begin{figure}[t!]
    \centering
    \begin{subfigure}[t]{0.44\linewidth}
        \centering
        \includegraphics[width=1\linewidth]{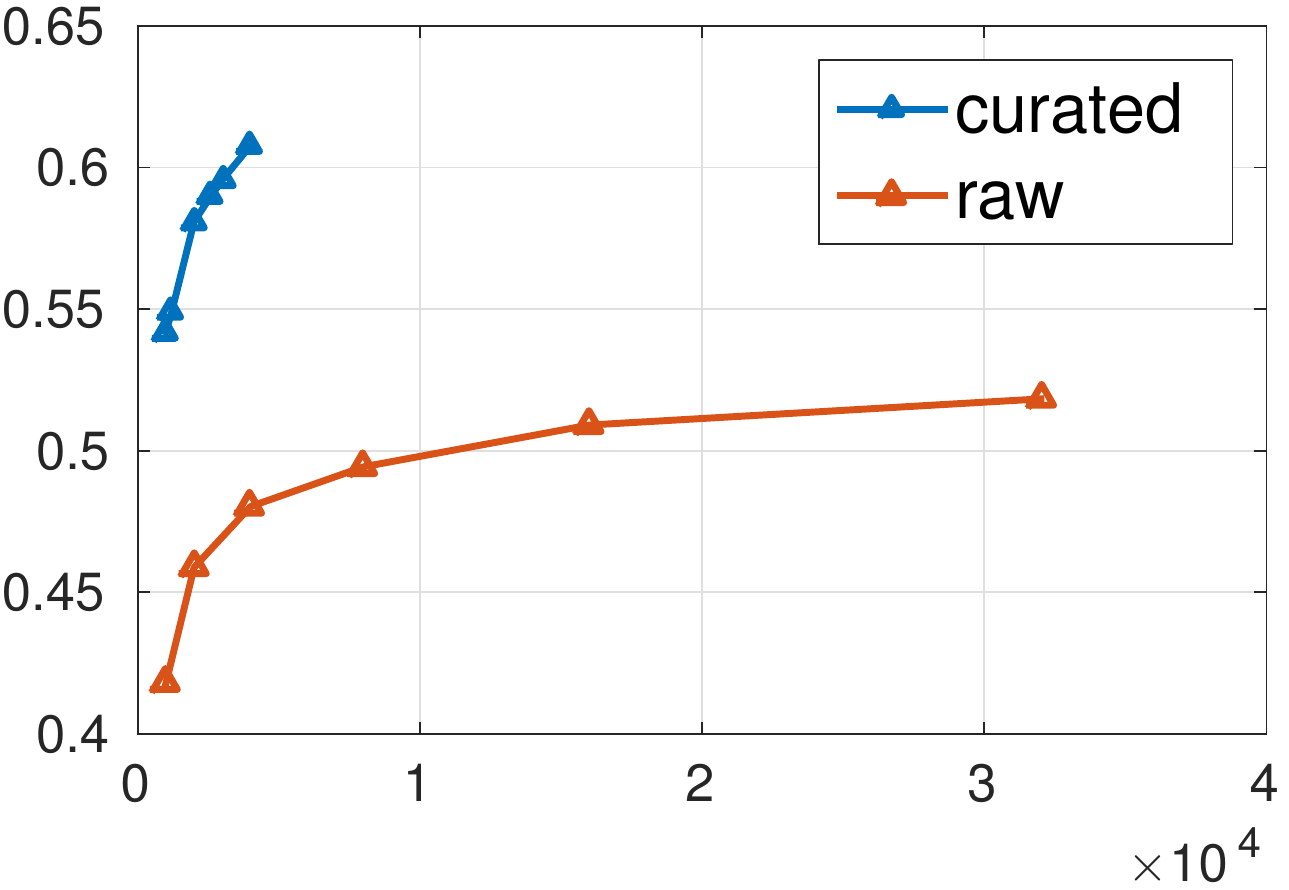}
        \caption{Testing on MV-40.}
    \end{subfigure}
    ~
    \begin{subfigure}[t]{0.44\linewidth}
        \centering
		\includegraphics[width=1\linewidth]{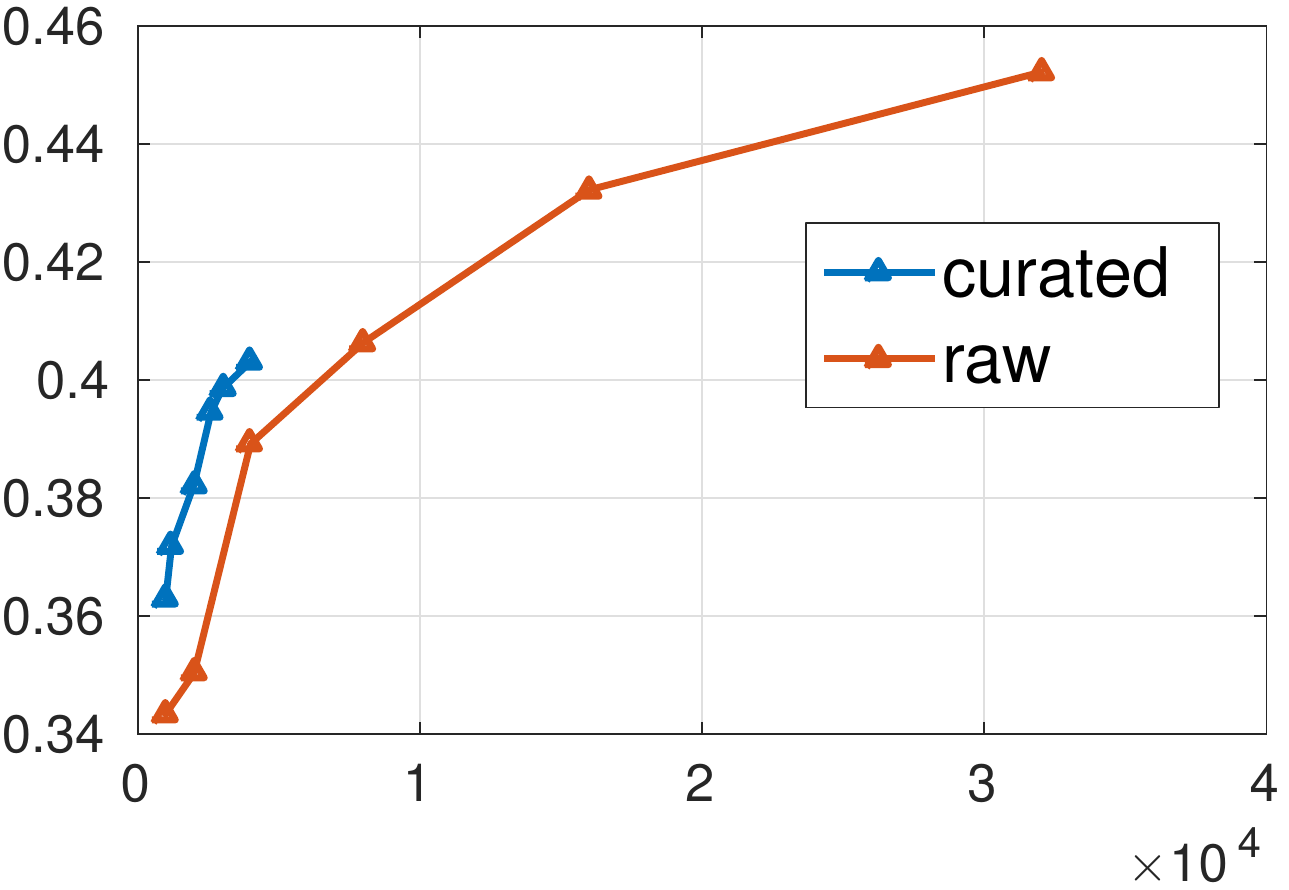}
		\caption{Testing on MV-58K.}
    \end{subfigure}%
    \vspace{-3mm}
    \caption{To evaluate the effect of data curation, we compare the accuracy of models trained and tested on the curated MV-40 versus uncurated data (obtained by targeted querying from the open-world for additional videos with the 40 given tags). The curated testset (a) more
    closely resembles existing recognition datasets while (b) is representative of tag prediction ``in the wild''. 
More training data improves action
    recognition (MV-40), but curated data is significantly more effective than
    a large, raw training set. For tag prediction (b), a modest amount
    of additional raw training data ($<$ 2X) rivals the accuracy of a manually
    curated training set. \vspace{-5mm}}
    \label{fig:clean}
\end{figure}

{\bf Curated vs raw data:} When comparing models trained on curated versus open-world data (Fig.~\ref{fig:clean}), we see that additional data always helps, but given a fixed amount of training data, curated data performs better. Curated data is
designed to mimic existing recognition datasets which use concise notions of
visual categories, while the raw dataset captures problem of tag-prediction
``in the wild''.  For example, videos of people talking about {\tt \small
\#archery} are removed in MV-40, but exist in MV-58k (Fig.~\ref{fig:visbreakdown}).  When
comparing accuracy on tag-prediction, we find that a clean dataset replicates
the accuracy of a 2X larger raw dataset.

\begin{figure}[t!]
\centering
\includegraphics[width=\linewidth]{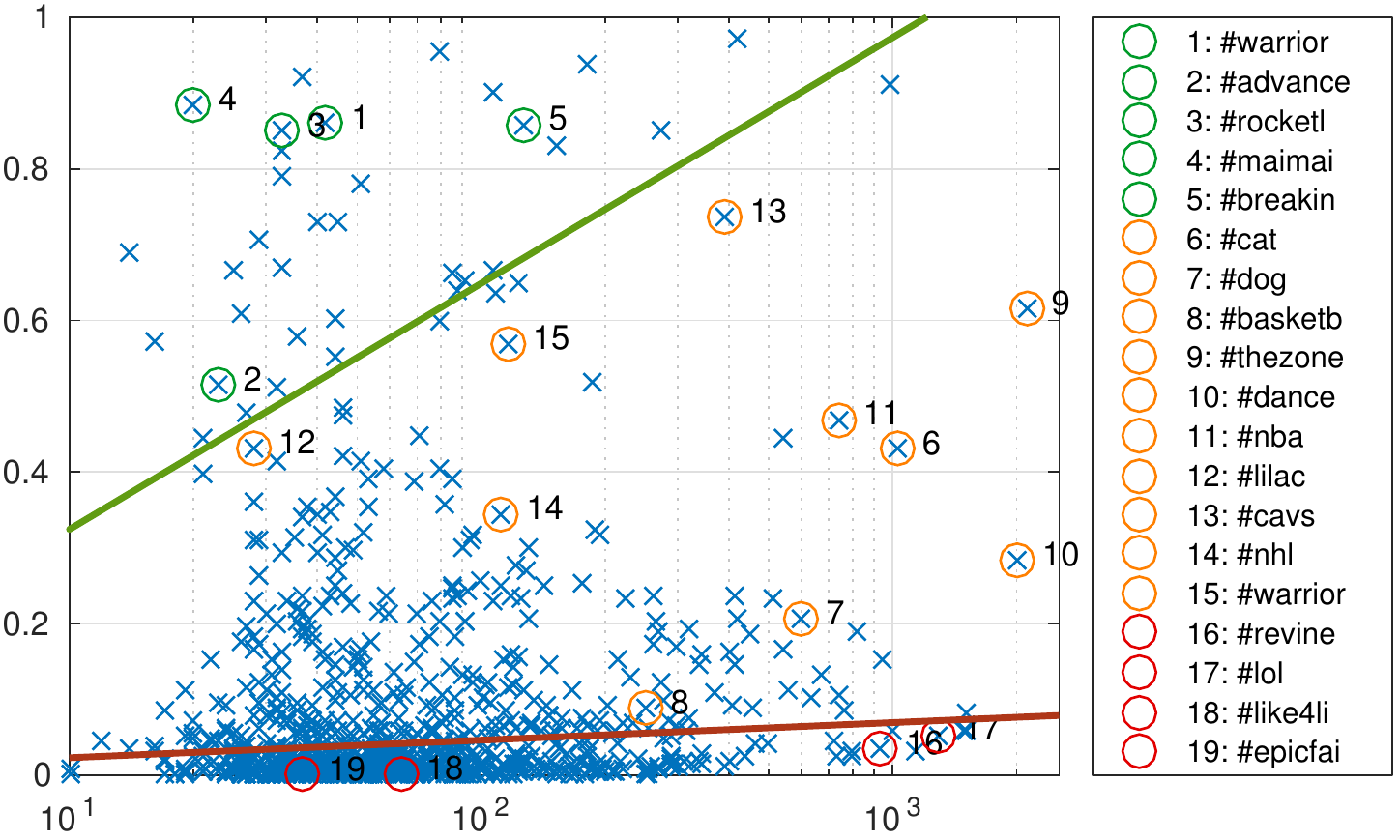}
\caption{Open-world tag prediction. The scatter plot shows per-tag APs vs (log) number of training examples. We rank the ``learnability'' of a tag by the ratio of its AP to (log) number of training examples, and draw lines to loosely denote regions of easy, challenging, and unlearnable tags. Unlearnable tags appear to correspond to ``stopwords'' such as {\tt \#revine,lol} that do not capture video content. The per-tag $mAP_{T}$ is 0.05.}
\label{fig:open}
\end{figure}

{\bf Closed vs open vocabularies:} We now move beyond our 40 selected tags to evaluation of an open-world vocabulary. Because the amount of training data variables considerably per tag, we plot test accuracy as a function of training-set size in Fig.~\ref{fig:open}. We tend to see different performance regimes. ``Easy'' tags perform well even with little training data, likely due to a characteristic appearance that is easy to learn from little training data  ({\tt \#cavs, \#warriors}).  ``Challenging'' tags appear to contain appearance variation, but are learnable with additional data ({\tt \#dogs, \tt \#soccer}). ``Unlearnable'' tags remain near-zero AP even given lots of training data ({\tt \#revine, \#lol}).  We posit that these can be treated as stopwords that fail to capture much semantic meaning of the video.

\begin{figure}[t!]
\begin{subfigure}[b]{.5\linewidth}
\centering
\raisebox{2.5\height}{
\begin{tabular}{c|ccc|}
Category & Non-causal & Causal & Adaptive\\
\hline
40 categories & 44.55\ & 38.05\ & 43.60\\
{\tt \small \#climbing} & 6.87\ & 5.27\ & 21.16\\
\end{tabular}}
\caption{}
\end{subfigure}
~
\begin{subfigure}[b]{.45\linewidth}
\centering
\includegraphics[width=\linewidth]{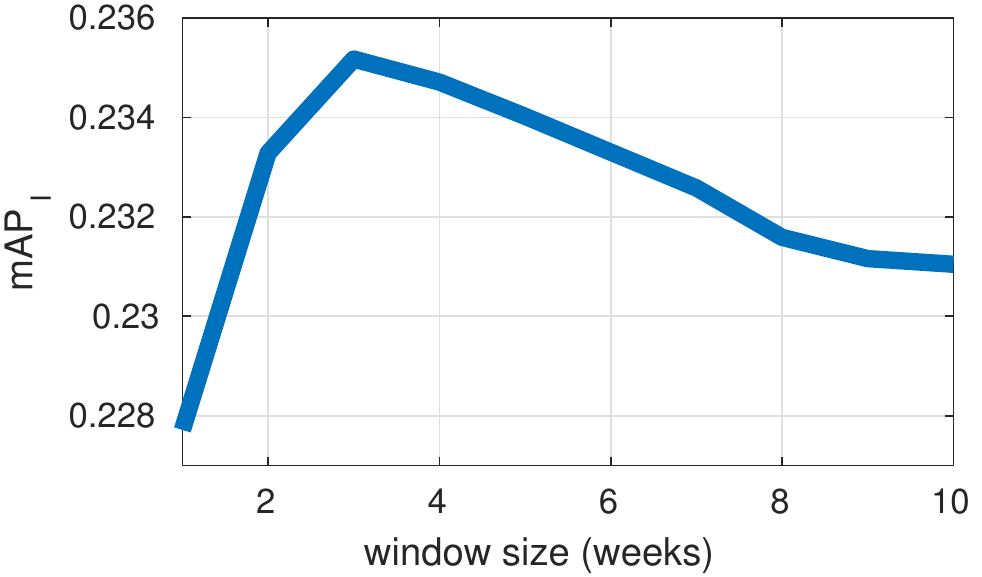}
\caption{}
\end{subfigure}
\vspace{-5pt}
\caption{Temporal dynamics. We plot performance of different temporal training strategies in ({\bf a}). Non-causal models perform the best but may be impractical because they must be trained on future videos.  Adaptive models trained on recent videos perform better than a fixed training set because tag topics tend to temporally evolve. In ({\bf b}), we analyze a hybrid causal-adaptive strategy that learns causal classifiers with adaptive Platt calibration (calibrated on the past $w$ weeks). With the window size at $w=3 $, the performance improves to 23.5\% (from 21\% with no calibration or calibrated on training data). If we calibrate on the ideal test distribution in the future, performance jumps significantly to 28\%, indicating that temporal models that can predict {\em future} popularity of tags would significantly improve accuracy.\label{fig:time}.}
\vspace{-10pt}
\end{figure}

{\bf Temporal dynamics:} We now examine the time-varying properties of
micro-video content and tags (Fig.~\ref{fig:time}). We refer the reader to the caption for a detailed analysis, but we find that causal prediction (where one only has access to data from the past) is much harder than the non-causal counterpart, suggesting that micro-videos are not ``iid'' over time.  Benchmarks that use iid resampling may over-estimate real-world performance on streaming data.  We find that much of this temporal variation can be explained by fluctuations of tag popularity, whose degree of variation can vary dramatically across classes.


{\bf Conclusion:} We introduce a open-world dataset of micro-videos, which lie in a regime between single images and typical videos, allowing for easy capture, storage, and processing. They contain micro-narratives captured from viewpoints typically not studied in computer vision. Because they are naturally diverse, pre-trimmed, and user-annotated, they can be used a live testbed for open-world evaluation of video understanding systems. We conclude with an intriguing thought: rather than distributing a fixed benchmark dataset (which historically leads to eventual overfitting~\cite{ponce2006dataset}), our analysis suggests that we can instead distribute a benchmark script that evaluates models on live open-world micro-videos. We think the time is right to consider video recognition out in-the-open!

\clearpage

\bibliography{egbib}
\end{document}